\documentclass[conference]{IEEEtran}
\IEEEoverridecommandlockouts
\usepackage{cite}
\usepackage{amsmath,amssymb,amsfonts}
\usepackage{algorithmic}
\usepackage{graphicx}
\usepackage{textcomp}
\usepackage{xcolor}
\usepackage{multirow}
\usepackage{enumitem}
\usepackage{booktabs}

\def\BibTeX{{\rm B\kern-.05em{\sc i\kern-.025em b}\kern-.08em
    T\kern-.1667em\lower.7ex\hbox{E}\kern-.125emX}}
\begin{document}

\title{LL4G: Self-Supervised Dynamic Optimization for Graph-Based Personality Detection
\thanks{This work was supported by the Alan Turing Institute and DSO National Laboratories under a grant on improving multimodal misinformation detection through affective analysis. Additional support was provided by the Interdisciplinary Research Pump-Priming Fund, University of Southampton.}
}

\author{
\IEEEauthorblockN{Lingzhi Shen}
\IEEEauthorblockA{
\textit{University of Southampton} \\
Southampton, United Kingdom \\
l.shen@soton.ac.uk}
\and
\IEEEauthorblockN{Yunfei Long}
\IEEEauthorblockA{
\textit{University of Essex} \\
Essex, United Kingdom \\
yl20051@essex.ac.uk}
\and
\IEEEauthorblockN{Xiaohao Cai}
\IEEEauthorblockA{
\textit{University of Southampton} \\
Southampton, United Kingdom \\
x.cai@soton.ac.uk}
\and
\IEEEauthorblockN{Guanming Chen}
\IEEEauthorblockA{
\textit{University of Southampton} \\
Southampton, United Kingdom \\
gc3n21@soton.ac.uk}
\and
\IEEEauthorblockN{Yuhan Wang}
\IEEEauthorblockA{
\textit{Birkbeck, University of London} \\
London, United Kingdom \\
ywang32@student.bbk.ac.uk}
\and
\IEEEauthorblockN{Imran Razzak}
\IEEEauthorblockA{
\textit{Mohamed bin Zayed University of Artificial Intelligence} \\
Abu Dhabi, United Arab Emirates \\
imran.razzak@mbzuai.ac.ae}
\and
\IEEEauthorblockN{Shoaib Jameel}
\IEEEauthorblockA{
\textit{University of Southampton} \\
Southampton, United Kingdom \\
M.S.Jameel@southampton.ac.uk}
}



\maketitle

\begin{abstract}
Graph-based personality detection constructs graph structures from textual data, particularly social media posts. Current methods often struggle with sparse or noisy data and rely on static graphs, limiting their ability to capture dynamic changes between nodes and relationships. This paper introduces LL4G, a self-supervised framework leveraging large language models (LLMs) to optimize graph neural networks (GNNs). LLMs extract rich semantic features to generate node representations and to infer explicit and implicit relationships. The graph structure adaptively adds nodes and edges based on input data, continuously optimizing itself. The GNN then uses these optimized representations for joint training on node reconstruction, edge prediction, and contrastive learning tasks. This integration of semantic and structural information generates robust personality profiles. Experimental results on Kaggle and Pandora datasets show LL4G outperforms state-of-the-art models.
\end{abstract}


\section{Introduction}
Personality comprises traits or qualities that shape an individual’s unique patterns of thinking, feeling, and behaving \cite{cervone2022personality}. The study of personality is imperative for understanding interpersonal relationships, personal behaviors, and preferences in various contexts, including education, workplaces, and social environments. Among the myriad theories of personality, the Myers-Briggs Type Indicator (MBTI) \cite{quenk2009essentials} is one of the most widely used. MBTI classifies individuals into 16 distinct personality types based on four opposing dimensions: Extroversion vs. Introversion (I/E), Sensing vs. Intuition (S/N), Thinking vs. Feeling (T/F), and Judging vs. Perceiving (J/P).

In recent years, an increasing number of researchers have employed machine learning algorithms to analyze textual inputs and numerical records for personality prediction applications. Social media data, in particular, have shown significant potential across various scenarios \cite{chen2017leveraging, shen2025gamed}. For example, a study leveraging BiLSTM with attention mechanisms achieved remarkable results in detecting psychopathy traits from social media text \cite{asghar2021detection}. Since the emergence of the Transformer architecture, it has revolutionized text-related tasks, with its effectiveness demonstrated in research combining sentiment lexicons with the BERT model for multi-label personality detection \cite{ren2021sentiment}. Additionally, various Transformer-based variants have frequently been adopted in hybrid approaches, such as combining them with ULMFiT \cite{halimawan2022bert} or fine-tuning for specific tasks \cite{alshouha2024personality}, to enhance recognition performance.

Graph-based personality detection has emerged as an important research direction. This approach leverages graph structures to model complex relationships among users, textual content, and personality traits, framing personality detection as a graph representation learning problem. Sophisticated methods, such as D-DGCN \cite{yang2023orders} and Semi-PerGCN \cite{zhu2024data}, utilize graph convolutional networks to model the structure of user-generated content. For instance, D-DGCN focuses on the dynamic relationships between posts, while Semi-PerGCN integrates psycholinguistic features with unsupervised learning. Moreover, hierarchical classifier models using Graph Attention Networks (GAT) have been applied to automatically detect personality traits from text \cite{roy2023personality}.

\begin{figure*}[htbp]
    \centering
    \includegraphics[scale=0.18]{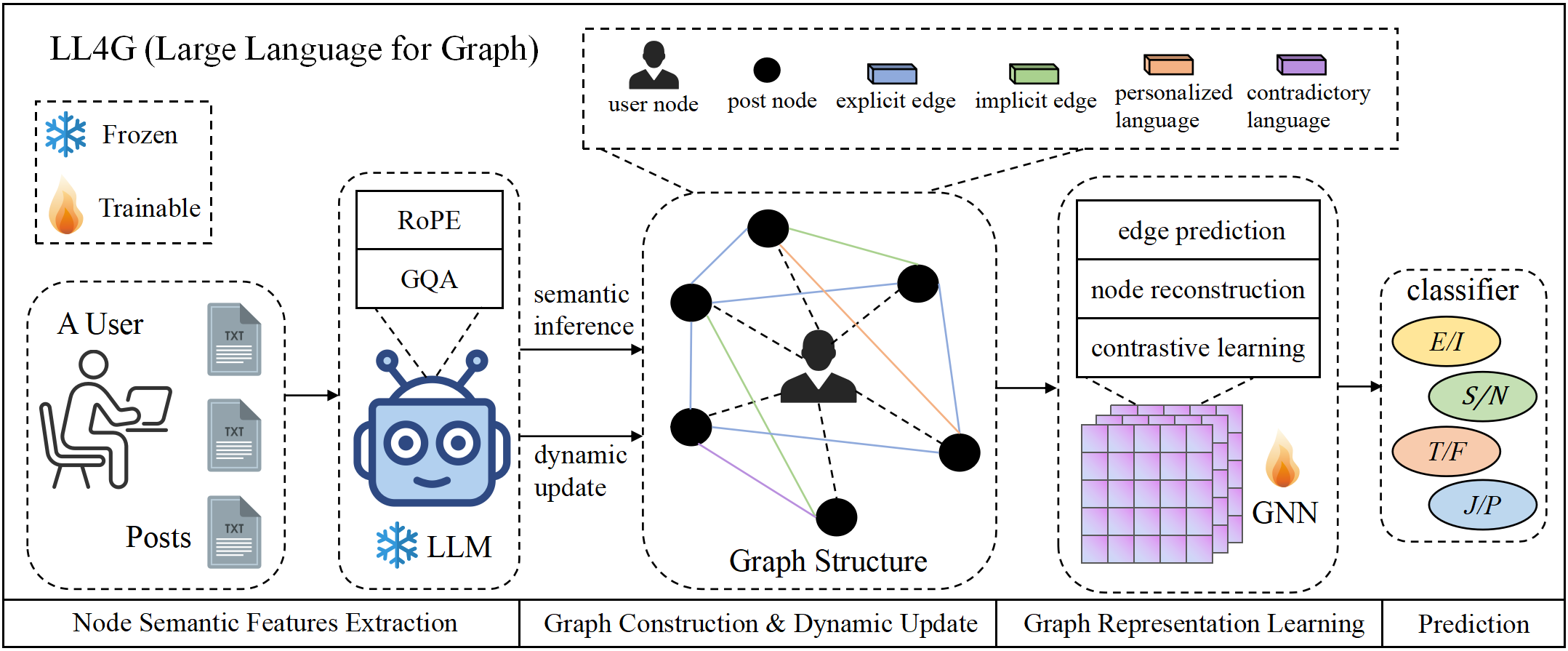} %
    \caption {The architecture diagram of the proposed LL4G. It illustrates the workflow pipeline, encompassing key stages such as node semantic extraction, graph construction and dynamic updates, graph representation learning, and personality prediction.}
    \label{fig:diagram}
    \vspace{-0.05in}
\end{figure*}

Personality detection poses unique challenges compared to other text classification tasks \cite{lu2024fact}. Datasets often suffer from severe class imbalances, while short social media posts lead to data sparsity, limiting feature extraction. Personality boundaries are also inherently ambiguous and context-dependent, with statements varying in interpretation based on context, adding noise and complicating predictions. Recently, large language models (LLMs) have shown promise in addressing these issues. By leveraging LLMs' advanced semantic, emotional, and linguistic insights, text augmentation techniques improve feature representation in smaller models \cite{hu2024llm}. LLMs excel at capturing complex contextual relationships and subtle linguistic cues, effectively inferring critical information that earlier models often miss.

In this study, we propose LL4G, a self-supervised framework that leverages LLMs to optimize graph structures for personality detection. LLMs extract deep semantic features and contextual dependencies from posts, generating high-quality node representations and constructing graphs based on semantic similarity and relational reasoning. A unique ``user node" aggregates information from all posts, providing essential contextual background. Unlike previous methods, LL4G incorporates explicit and implicit relational inference to simulate complex interactions, emphasizing personalized and contradictory linguistic features critical for distinguishing traits and capturing nuanced behaviors. Its dynamic adaptability allows real-time graph updates as user data evolves, with a self-supervised mechanism inferring and completing new nodes and edges to address static graph limitations. A GNN performs representation learning on the dynamic graph, optimized through multi-task training, including node reconstruction, edge prediction, and contrastive learning. This continuous process enhances LL4G’s generalization capabilities, ultimately producing a holistic personality profile by integrating semantic and structural features.

\noindent \textbf{Contributions:} 
We propose a dynamic graph optimization method that updates graph structures through self-supervised learning, addressing the limitations of static graph approaches and enhancing representation for personality detection. We introduce an adaptive graph construction method that extracts high-quality semantic features from personality-related texts, mitigating feature sparsity and noise. Our model LL4G outperforms state-of-the-art methods on benchmark datasets.

\section{Methodology}
LL4G is a novel self-supervised architecture that leverages LLMs' advanced natural language understanding to enhance graph-based personality detection by integrating semantic and structural features, see Fig.~\ref{fig:diagram}. 

\subsection{Node Semantic Features Extraction}  
Llama 3 is used to extract high-quality semantic features from each user's posts, leveraging its advanced architectures, Grouped Query Attention (GQA) and Rotary Position Embeddings (RoPE). RoPE encodes relative positional information into the dot product of attention, where a position \( p \) in the sequence \( x \) is transformed using a rotation matrix, i.e., 
\[
x_p = R(p)x, \quad R(p) = 
\begin{bmatrix} 
\cos(p) & -\sin(p) \\ 
\sin(p) & \cos(p) 
\end{bmatrix}.
\]  

GQA enhances computational efficiency by dividing queries, keys, and values into \( G \) groups. Attention for each group is computed as:  
\[
\text{Attention}(Q, K, V) = \text{softmax}\left(\frac{QK^\top}{\sqrt{d_k / G}}\right)V,
\]  
where \( d_k \) is the dimensionality of the keys. Outputs from all groups are concatenated, reducing redundancy while preserving semantic precision.

Using these mechanisms, each post \( T_i \) is transformed into a semantic embedding \( h_i = \text{Llama}(T_i) \in \mathbb{R}^d \). The embeddings are aggregated into the node feature matrix \( H = [h_1, h_2, \cdots, h_N]^\top \in \mathbb{R}^{N \times d} \), where \( N \) is the total number of posts per user. This matrix serves as the input for graph representation learning. 

\subsection{Graph Construction and Dynamic Update}
We leverage Llama's semantic inference capabilities to model relationships across a user's posts, constructing a graph with both explicit and implicit edges.

Explicit edges are generated based on the semantic similarity and personalization between posts. The edge weight \( A_{ij}^{\text{explicit}} \) between two posts \( T_i \) and \( T_j \) is defined as:
\[
    A_{ij}^{\text{explicit}} = \text{sim}(h_i, h_j) + \lambda_p \cdot P_{ij}, \quad \text{sim}(h_i, h_j) = \frac{h_i^\top h_j}{\|h_i\| \|h_j\|} ,
\]
\noindent where \( \text{sim}(h_i, h_j) \) is the cosine similarity measuring the semantic proximity between $h_i$ and $h_j$, \( P_{ij} \) quantifies the relationship between the two posts in terms of personalized language (Personalized language reflects an individual's unique linguistic style), and \( \lambda_p \) is a weighting factor that controls the contribution of personalized language in the explicit edge construction. The adjacency matrix for explicit edges is
\[
    A^{\text{explicit}}_{ij} =
    \begin{cases}
    A_{ij}^{\text{explicit}}, & \text{if } \text{sim}(h_i, h_j) + \lambda_p \cdot P_{ij} > \tau, \\
    0, & \text{otherwise};
    \end{cases}
\]
\noindent \( \tau \) is the similarity threshold for retaining high-quality edges.

Implicit edges capture deeper latent relationships, such as contradictory language similarity, which identifies inconsistencies in expressed opinions across a user's different posts. Llama performs contextual reasoning to generate implicit edge weights as:
\[
    A_{ij}^{\text{implicit}} = f_{\text{Llama}}(T_i, T_j) + \lambda_c \cdot C_{ij} ,
\]
\noindent where \( f_{\text{Llama}}(T_i, T_j) \) is a score function quantifying the implicit relationship between posts \( T_i \) and \( T_j \), \( C_{ij} \) quantifies the contradictory language relationship between the two posts, and \( \lambda_c \) is a weighting factor controlling the impact of contradictory language in the implicit edge construction. The adjacency matrix for implicit edges is defined as:
\[
A_{ij}^{\text{implicit}} =
\begin{cases}
A_{ij}^{\text{implicit}}, & \text{if } f_{\text{Llama}}(T_i, T_j) + \lambda_c \cdot C_{ij} > \tau', \\
0, & \text{otherwise};
\end{cases}
\]
\noindent \( \tau' \) is the threshold for the implicit relationship score. The final adjacency matrix combining explicit and implicit edges is 
\[ A = A^{\text{explicit}} + A^{\text{implicit}}.
\]

To enhance the graph structure, we introduce a user node \( u \) that represents the user as a whole and connects to all post nodes. This user node aggregates information from all posts, providing a unified representation that complements individual post-level embeddings. The connection between the user node and each post node is established with the following weight: \( A_{u, i} = \text{user\_link}(u, h_i) \), where \( \text{user\_link}(u, h_i) \) quantifies the relationship between the user node \( u \) and the post node \( h_i \). 

The user node is added to the graph by extending the matrix \( A \) to include its connections with all posts, i.e.,
\[
A = \begin{bmatrix}
A^{\text{post}} & A^{\text{user}} \\
(A^{\text{user}})^\top & 0
\end{bmatrix} ,
\]
where \( A^{\text{post}} \) is the adjacency matrix representing relationships among post nodes, including both explicit and implicit edges, and \( A^{\text{user}} \) represents the connection weights between the user node and the post nodes. The user node’s embedding \( h_u \) is generated by aggregating the embeddings of all connected post nodes using an attention mechanism. The aggregation is defined as:
\[
h_u = \sum_{i=1}^N \alpha_i h_i, \quad \alpha_i = \frac{\exp(w^\top h_i)}{\sum_{j=1}^N \exp(w^\top h_j)},
\]
\noindent where \( w \) is a learnable weight vector that determines the importance of each post \( h_i \) in contributing to the user representation \( h_u \), and \( \alpha_i \) is the corresponding attention weight. The inclusion of the user node ensures that the graph not only captures post-level interactions but also reflects the user’s holistic characteristics.

In the dynamic graph update mechanism, when a new post \( T_{N+1} \) is added, relationships with existing nodes are recalculated to update the adjacency matrix \( A \), i.e., \( A_{i, N+1} = \text{sim}(h_i, h_{N+1}) + \lambda_p \cdot P_{i, N+1} \) and \( A_{N+1, j} = f_{\text{Llama}}(T_{N+1}, T_j) + \lambda_c \cdot C_{N+1, j} \), with \( A_{i, N+1} \) and \( A_{N+1, j} \) represent the explicit and implicit edges between the new post \( T_{N+1} \) and existing posts \( i, j \), respectively. Additionally, the connection between the user node \( u \) and the new post \( T_{N+1} \) is updated as \( A_{u, N+1} = \text{user\_link}(u, h_{N+1}) \), quantifying the relationship between the user node and the new post based on semantic similarity or metadata. The adjacency matrix \( A \) is extended to accommodate the new post \( T_{N+1} \) and its connections with all existing nodes, including the user node.

\subsection{Graph Representation Learning}
After constructing the graph structure, we select DGCNs as a representative of GNNs for graph representation learning. DGCNs capture high-order neighbourhood information and effectively adapt to large-scale graphs, making them more suitable for dynamic optimization objectives.

The DGCN propagates messages iteratively across nodes, allowing each post and the user node to aggregate semantic and structural information from neighbours. Recall $A$ denotes the adjacency matrix of the graph (comprising both explicit and implicit edges, as well as connections to the user node), and let $H^{(0)}$ be the initial node feature matrix extracted from Llama embeddings. The GCN updates the node representations layer-by-layer using the following propagation rule, i.e.,
\[
H^{(l+1)} = \sigma\left(\tilde{D}^{-1/2} \tilde{A} \tilde{D}^{-1/2} H^{(l)} W^{(l)}\right),
\] 
where $\tilde{A} = A + I$ is the adjacency matrix augmented with self-loops, $\tilde{D}$ is its degree matrix, $W^{(l)}$ is the learnable weight matrix at layer $l$, and $\sigma(\cdot)$ is the activation function. This process ensures that each node representation integrates semantic similarity and structural dependencies among posts, as well as the influence of the user node.

To optimize the representations of posts and user embeddings, we jointly train the model on node reconstruction, edge prediction, and contrastive learning tasks. Among these, node reconstruction and edge prediction serve as the core objectives, while contrastive learning acts as an auxiliary task. 

The node reconstruction task preserves the semantic integrity of the original input features extracted from Llama in the final embeddings. For a node $i$, the input embedding $h_i^{\text{input}}$ is compared to its final representation $h_i^{\text{output}}$, produced by the DGCN, by minimizing the Euclidean distance
\[
    L_{\text{node}} = \frac{1}{N+1} \sum_{i=1}^{N+1} \| h_i^{\text{output}} - h_i^{\text{input}} \|_2^2 ,
\]
where the user node \( h_u \) is included in the computation as part of graph. This objective preserves the critical semantic information throughout the propagation layers of the DGCN.

The edge prediction task focuses on modelling the relationships between nodes by predicting the existence of edges in the graph. For each pair of nodes $i$ and $j$, the edge likelihood is predicted as $\sigma(h_i^\top h_j)$, where $\sigma(x) = 1 / (1 + e^{-x})$ is the sigmoid function. The edge prediction loss is then defined as:
\[
L_{\text{edge}} = - \sum_{(i,j) \in E} \log \sigma(h_i^\top h_j) - \sum_{(i,j) \notin E} \log \left(1 - \sigma(h_i^\top h_j) \right) ,
\]
\noindent where $E$ is the set of observed edges in the graph, and $(i, j) \notin E$ represents negative samples. This objective ensures that the model learns to accurately represent both explicit and implicit relationships between nodes.

To enhance the discriminative power of node embeddings, we incorporate contrastive learning as an auxiliary objective, encouraging the model to structure the embedding space such that semantically similar nodes are drawn closer, while dissimilar ones are pushed apart. For a node $i$, the positive sample \( h_i^+ \) is a neighbouring node in the graph, while the negative samples \( \mathcal{N}^- \) are sampled from unrelated nodes. The contrastive loss is defined as:
\[
L_{\text{contrastive}} = - \frac{1}{N+1} \sum_{i=1}^{N+1} \log \frac{s_{i,i^+}}{s_{i,i^+} + \sum_{j \in \mathcal{N}^-} s_{i,j}} ,
\]
where \( s_{i,j} = \exp(h_i^\top h_j / \tau) \) is the similarity score between nodes \( i \) and \( j \), and \( \tau \) is a temperature parameter that controls the concentration of the distribution. 

Three tasks are integrated into a single optimization objective, with weighting coefficients for balancing, i.e.,
\[ L_{\text{total}} = \alpha L_{\text{node}} + \beta L_{\text{edge}} + \gamma L_{\text{contrastive}},
\]
where $\alpha$, $\beta$, and $\gamma$ are hyperparameters that determine the relative importance of each task. Since contrastive learning is auxiliary, $\gamma$ is set to a smaller value to ensure it does not dominate the optimization process. By jointly optimizing these tasks, the framework achieves a balance between preserving individual node features, accurately modelling relationships within the graph, and ensuring discriminative representations. 

\subsection{Personality Prediction}
The final representation of the user node \( h_u \), learned through DGCN, encapsulates both the semantic richness of individual posts and their structural relationships, and serves as a comprehensive representation of the user's personality profile.


Each MBTI dimension (e.g., Extroversion vs. Introversion) is treated as a binary classification task. Final classification supervision uses cross-entropy loss, i.e.,
\[
L_{\text{class}} = - \frac{1}{M} \sum_{k=1}^M \sum_{c=1}^2 y_k^{(c)} \log \hat{y}_k^{(c)},
\]
where \( M \) is the total number of users, \( y_k^{(c)} \) is the ground-truth label for user \( k \) in class \( c \), and \( \hat{y}_k^{(c)} \) is the predicted probability. The overall loss \( L_{\text{final}} \) for the framework combines the classification objective \( L_{\text{class}} \) with the previously introduced self-supervised training tasks \( L_{\text{total}} \). The complete loss is defined as \( L_{\text{final}} = L_{\text{class}} + \lambda L_{\text{total}} \), where \( \lambda \) is a hyperparameter that controls the contribution of the self-supervised tasks to the overall optimization.

\begin{table*}[htbp]
    \caption{Performance comparison in terms of the Macro-F1 scores (\%) across four dimensions and the overall average (Avg). The results are evaluated between our LL4G and state-of-the-art baselines on the Kaggle and Pandora datasets.}
    \centering
    \scalebox{0.96}
    {
    \begin{tabular}{l|cccc|c|cccc|c}
        \hline
        \multirow{2}{*}{\textbf{Methods}} & \multicolumn{5}{c|}{\textbf{Kaggle}} & \multicolumn{5}{c}{\textbf{Pandora}} \\
        \cline{2-11}
        & E/I & S/N & T/F & J/P & \textbf{Avg} & E/I & S/N & T/F & J/P & \textbf{Avg} \\
        \hline
        BiLSTM+CNN \cite{sun2018personality} & 60.82 & 59.51 & 67.48 & 58.76 & 61.64 & 53.42 & 50.89 & 60.35 & 52.47 & 54.78 \\
        BERT \cite{keh2019myers} & 64.65 & 57.12 & 77.95 & 65.25 & 66.24 & 54.22 & 48.71 & 64.70 & 56.07 & 56.56 \\
        RoBERTa \cite{liu2019roberta} & 61.89 & 57.59 & 78.69 & 70.07 & 67.06 & 54.80 & 55.12 & 63.78 & 55.94 & 57.41 \\
        AttRCNN \cite{xue2018deep} & 59.74 & 64.08 & 78.77 & 66.44 & 67.25 & 48.55 & 56.19 & 64.39 & 57.26 & 56.60 \\
        SN+Attn \cite{lynn2020hierarchical} & 65.43 & 62.15 & 78.05 & 63.92 & 67.39 & 56.98 & 54.78 & 60.95 & 54.81 & 56.88 \\
        Transformer-MD \cite{yang2021multi} & 66.08 & 69.10 & 79.19 & 67.50 & 70.47 & 55.26 & 58.77 & 69.26 & 60.90 & 61.05 \\
        PQ-Net \cite{yang2021learning} & 68.94 & 67.65 & 79.12 & 69.57 & 71.32 & 57.07 & 55.26 & 65.64 & 58.74 & 59.18 \\
        TrigNet \cite{yang2021psycholinguistic} & 69.54 & 67.17 & 79.06 & 67.69 & 70.86 & 56.69 & 55.57 & 66.38 & 57.27 & 58.98 \\
        PsyCoT \cite{yang2023psycot} & 66.56 & 61.70 & 74.80 & 57.83 & 65.22 & 60.91 & 57.12 & 66.45 & 53.34 & 59.45 \\
        DEN \cite{zhu2024enhancing} & 69.95 & 66.39 & 80.65 & 69.02 & 71.50 & 60.86 & 57.74 & \textbf{71.64} & 59.17 & 62.35 \\
        PS-GCN \cite{liu2024ps} & 70.52 & 65.73 & 70.51 & 67.13 & 68.47 & 59.12 & 54.88 & 67.35 & 58.62 & 59.49 \\
        D-DGCN \cite{yang2023orders} & 69.52 & 67.19 & 80.53 & 68.16 & 71.35 & 59.98 & 55.52 & 70.53 & 59.56 & 61.40 \\
        TAE \cite{hu2024llm} & 70.90 & 66.21 & 81.17 & 70.20 & 72.07 & 62.57 & 61.01 & 69.28 & 59.34 & 63.05 \\
        \hline
        \textbf{LL4G} & \textbf{79.77} & \textbf{78.90} & \textbf{85.84} & \textbf{77.66} & \textbf{80.54} & \textbf{67.87} & \textbf{66.07} & 71.55 & \textbf{65.93} & \textbf{67.85} \\
        \hline
    \end{tabular}
    }
    \label{tab:macro-f1}
    \vspace{-0.05in}
\end{table*}

\section{Experiments}
\subsection{Experimental Setup}
\noindent \textbf{Datasets:}
We use the Kaggle and Pandora datasets, i.e., the most recognized benchmarks in MBTI personality detection. The Kaggle dataset, sourced from the PersonalityCafe forum, contains posts from 8,675 users, with each contributing 45--50 posts spanning topics like psychology and personal experiences, showcasing diverse linguistic styles. Pandora, a larger and more varied corpus from Reddit, includes MBTI labels for 9,084 users with post counts ranging from dozens to hundreds, covering topics such as entertainment and technology. 

\noindent \textbf{Implementation Details:}
We leverage the ``Meta-Llama-3-8B-Instruct'' model, standardizing its output vector to a length of 4096 while keeping Llama frozen. 
We use DGCNs for graph representation learning, the Adam optimizer with a learning rate of $1\times10^{-3}$, and a dropout rate of 0.2 to prevent overfitting. The loss function is cross-entropy loss, and datasets are split into training, validation, and test sets (60\%, 20\%, and 20\%, respectively). During preprocessing, words matching personality labels are removed to prevent information leakage. Reported results are averages from 10 independent runs.

\noindent \textbf{Baselines and Evaluation Metrics:}
Table~\ref{tab:macro-f1} presents a comparison against various recent and relevant strong baseline selections. We used Macro-F1 (\%) scores as the evaluation metric that has been adopted in the baseline works.

\subsection{Overall Results}

Table~\ref{tab:macro-f1} compares LL4G with baselines, demonstrating an average improvement of 8.47\% over TAE on the Kaggle dataset and 4.80\% on Pandora. LL4G significantly outperforms all baselines across individual dimensions, except for a slight lag behind DEN on T/F in Pandora, and effectively reduces the performance gap between T/F and the other dimensions (E/I, S/N, J/P), showcasing its balanced prediction capability. Compared to other graph-based models like TrigNet, PS-GCN, DEN, and D-DGCN, LL4G achieves substantial gains, improving by 9.68\%, 12.07\%, 9.04\% and 9.19\% on Kaggle, and 8.87\%, 8.36\%, 5.50\% and 6.45\% on Pandora, respectively, consistently delivering superior results across all dimensions. These improvements highlight the design advantages of LL4G in capturing latent relationships within the data. By integrating LLM-enhanced semantic and structural features of graphs, LL4G dynamically adapts to the inherent complexities of diverse datasets and personality dimensions.

\subsection{Ablation Study}

\begin{table*}[htbp]
    \caption{The ablation study assessing the impact of different components on LL4G. Experiments were conducted on the Kaggle and Pandora datasets in terms of the evaluation metric Macro-F1 (\%) scores across the four dimensions and their average (Avg).}
    \centering
    \scalebox{0.96}
    {
    \begin{tabular}{l|cccc|c|cccc|c}
        \hline
        \multirow{2}{*}{\textbf{Components}} & \multicolumn{5}{c|}{\textbf{Kaggle}} & \multicolumn{5}{c}{\textbf{Pandora}} \\
        \cline{2-11}
        & E/I & S/N & T/F & J/P & \textbf{Avg} & E/I & S/N & T/F & J/P & \textbf{Avg} \\
        \hline
        Fully Fine-tuned Llama & 67.91 & 62.30 & 80.41 & 67.05 & 69.42 & 64.06 & 59.72 & 69.50 & 64.23 & 64.38 \\
        w/o Personalized Language & 79.56 & 76.07 & 83.37 & 76.51 & 78.88 & 64.50 & 59.98 & 69.29 & 65.47 & 64.81 \\
        w/o Contradictory Language & 79.62 & 77.56 & 85.78 & 76.95 & 79.98 & 63.84 & 65.64 & 69.73 & 64.14 & 65.84 \\
        w/o User Nodes & 79.40 & 77.23 & 84.41 & 75.61 & 79.16 & 63.84 & 65.13 & 70.06 & 62.86 & 65.47 \\
        w/o Node Reconstruction & 79.41 & 75.17 & 84.88 & 77.38 & 79.21 & 65.43 & 65.86 & 70.59 & 64.23 & 66.53 \\
        w/o Edge Prediction & 79.04 & 75.69 & 84.13 & 74.75 & 78.40 & 64.43 & 64.33 & 66.99 & 62.32 & 64.52 \\
        w/o Contrastive Learning & 77.82 & 74.85 & 83.42 & 76.79 & 78.22 & 65.97 & 65.26 & 70.83 & 64.81 & 66.72 \\
        Replace Llama with BERT & 74.45 & 70.81 & 80.46 & 60.91 & 71.66 & 63.26 & 53.49 & 68.80 & 61.37 & 61.73 \\
        Replace Llama with GPT-4 & 79.05 & 78.27 & \textbf{86.03} & 76.91 & 80.06 & 66.24 & \textbf{67.55} & 71.30 & \textbf{66.91} & \textbf{68.00} \\
        \hline
        \textbf{LL4G} & \textbf{79.77} & \textbf{78.90} & 85.84 & \textbf{77.66} & \textbf{80.54} & \textbf{67.87} & 66.07 & \textbf{71.55} & 65.93 & 67.85 \\
        \hline
    \end{tabular}
    }
    \label{tab:ablation}
    \vspace{-0.05in}
\end{table*}

\noindent \textbf{Feature Extractor:}
We evaluated various feature extractors in Table \ref{tab:ablation} for post semantics, noting a significant performance gap between BERT and Llama. LLMs like Llama excel in capturing rich contextual relationships, modelling long-range dependencies, and generating high-dimensional embeddings due to their large-scale parameters and advanced mechanisms such as RoPE and GQA. Both Llama and GPT-4 showed competitive performance: Llama slightly outperformed GPT on Kaggle, while GPT excelled on Pandora, likely due to Pandora's greater complexity and linguistic variability, better addressed by GPT's broader training corpus. Testing Llama 3 in a fully fine-tuned setting without the graph structure resulted in average performance drop of 11.12\% on Kaggle and 3.47\% on Pandora, underscoring the importance of graph structures in boosting predictive accuracy.

\noindent \textbf{Graph Construction:}
We evaluated our design in Table \ref{tab:ablation}  by removing the modelling of personalized language, contradictory language, and user nodes to validate the inspiration behind our framework. Results showed that the absence of these three components led to varying degrees of performance degradation in LL4G. The removal of personalized language had the greatest impact, as it often represents the most distinctive and expressive aspect of an individual's personality. The performance drop caused by the lack of contradictory language indicates the presence of underlying contradictions in users' post stances, a critical aspect that has been overlooked in earlier studies. Additionally, the inclusion of user nodes significantly contributes to capturing the holistic interactions among posts, further enhancing the overall performance. 

\noindent \textbf{Graph Representation Learning:}
We also conducted ablation studies in Table \ref{tab:ablation}  on node reconstruction, edge prediction, and contrastive learning to identify their distinct impacts on graph quality. The absence of edge prediction severely impacts LL4G's performance, leading to a significant 3.33\% drop in average performance on the Pandora dataset. This is because edge prediction is critical for modelling relationships between posts, further reinforcing the structural representations of the graph enhanced by the LLM. Similarly, the lack of node reconstruction and contrastive learning results in corresponding declines in overall performance, as these components are crucial for preserving the inherent semantics and ensuring the discriminability of embeddings within the graph.

\subsection{Case Study}

\begin{figure}[htbp]
    \centering
    \includegraphics[scale=0.08]{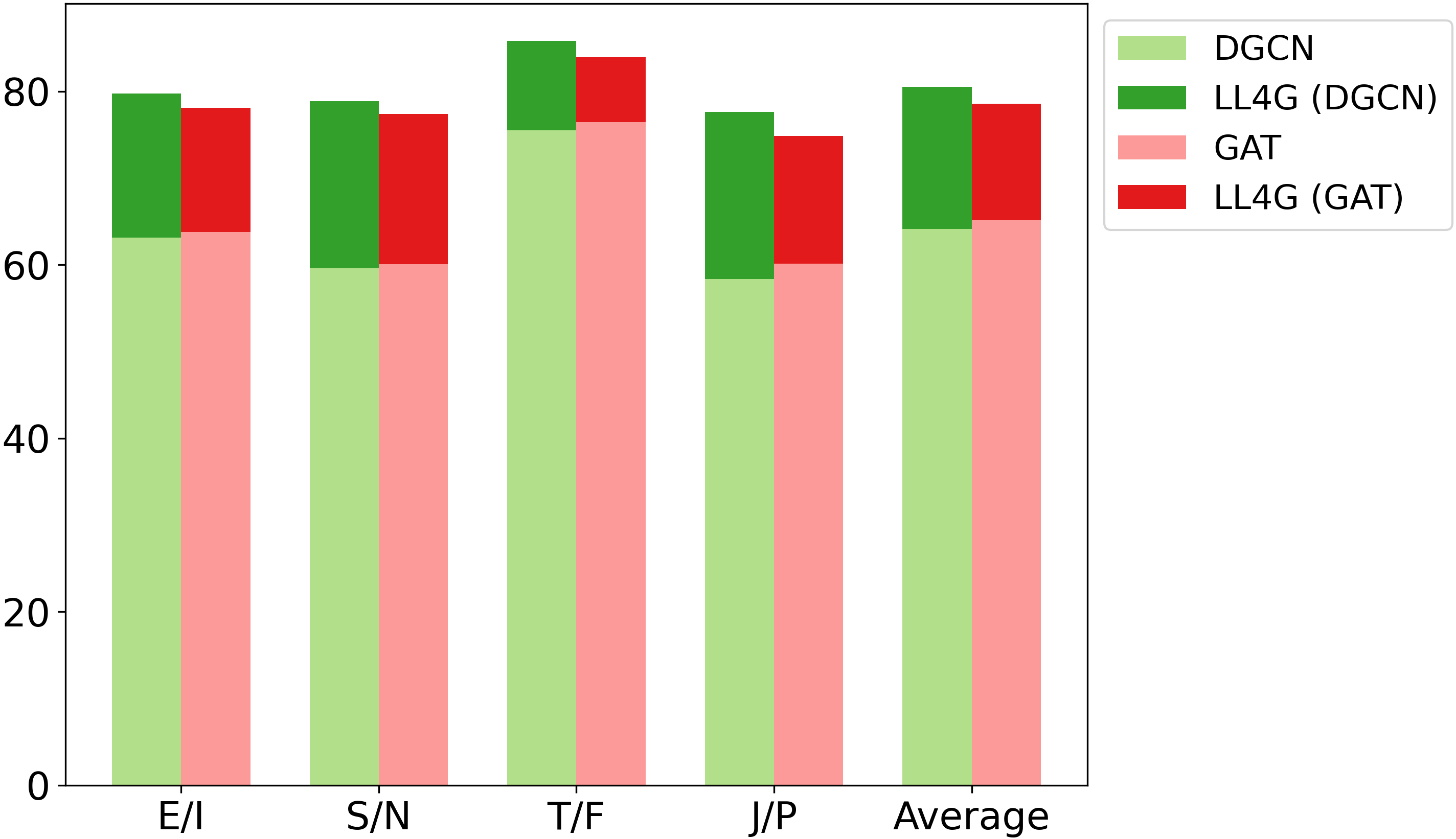} %
    \caption {Performance comparison between the LL4G framework and various graph-based models across four dimensions and their overall average in the form of the stacked bar chart in terms of Macro-F1 (\%) on the Kaggle dataset.}
    \label{fig:bar}
\end{figure}

\begin{figure}[htbp]
    \centering
    \includegraphics[scale=0.08]{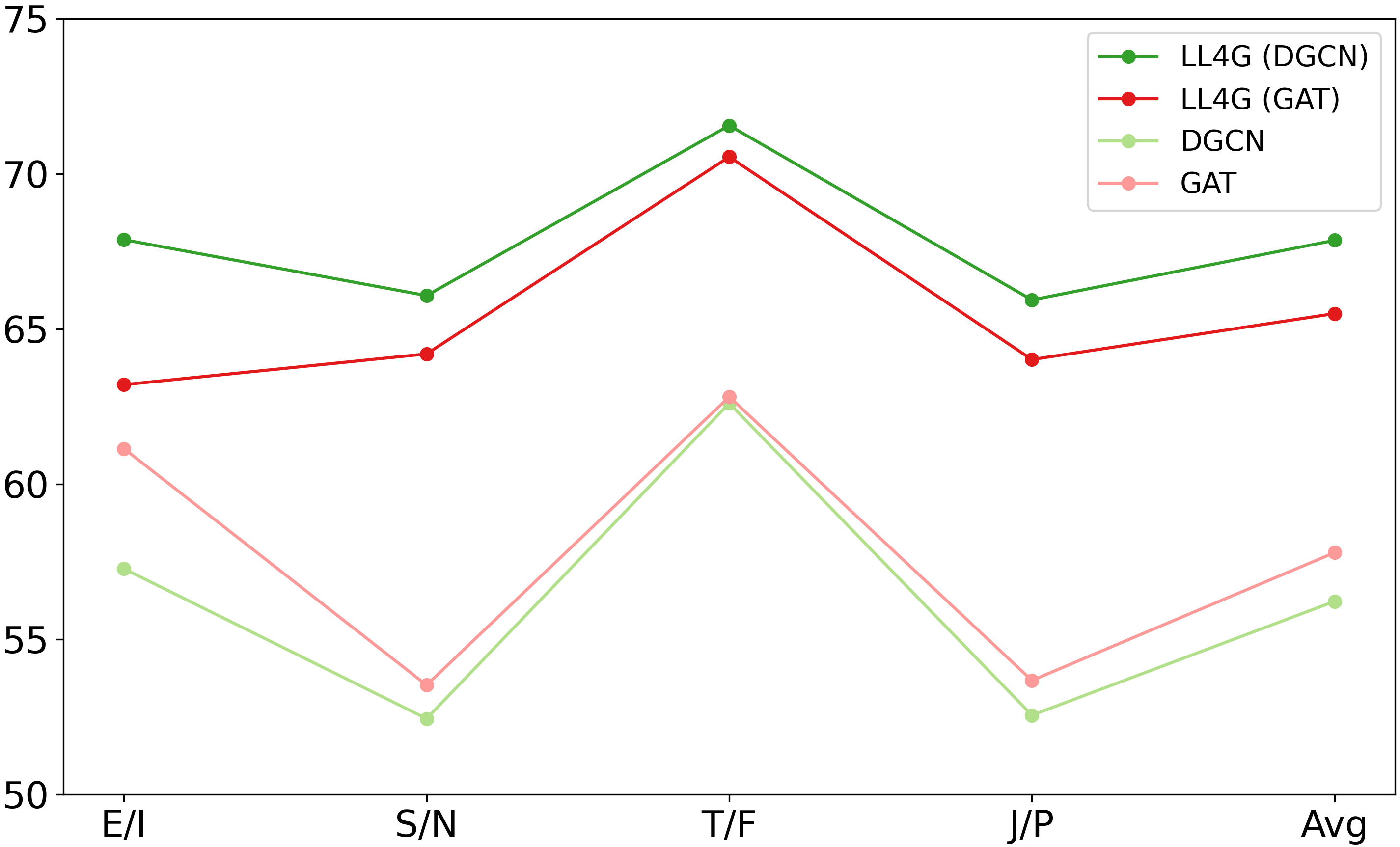} %
    \caption {Performance comparison bewteen the LL4G framework and various graph-based models across four dimensions and their overall average (Avg) in the form of line chart in terms of Macro-F1 (\%) on the Pandora dataset.}
    \label{fig:line}
    \vspace{-0.05in}
\end{figure}

To intuitively demonstrate LL4G's optimization effects on graph models, we visualize the performance of LL4G alongside standalone graph models such as DGCN and GAT. 
Fig.~\ref{fig:bar} and \ref{fig:line} demonstrate significant performance improvements in graph models achieved through the LL4G framework, highlighting how our novel design enhances prediction performance by improving graph structure quality.

We observed an intriguing phenomenon: standalone DGCN slightly underperformed compared to GAT on both datasets. However, with the enhancements provided by LL4G, DGCN outperformed GAT. This can be attributed to LL4G's LLM-driven graph optimizations, such as high-quality semantic and contextual extraction, which align particularly well with DGCN's characteristics. DGCN benefits significantly from LL4G's structure-level enhancements, as its deep graph representation learning effectively leverages high-order relationships and fine-grained semantics provided by the optimized graph. In contrast, GAT's reliance on attention mechanisms to assign weights limits the impact of additional enhancements. These findings demonstrate that LL4G's dynamic optimizations are especially effective for models like DGCN, which depend heavily on structural robustness.

\section{Conclusions}
We proposed LL4G -- a novel self-supervised framework for dynamically optimizing graph-based personality detection. LL4G leverages LLMs to extract deep semantic features and generate adaptive graph structures, followed by enhanced graph representation learning through joint training on node reconstruction, edge prediction, and contrastive learning. 
Experiments on Kaggle and Pandora benchmark datasets demonstrate that LL4G outperforms state-of-the-art models, with ablation studies confirming its effectiveness and robustness. Future work will explore multimodal extensions, e.g., integrating images and audio to enrich contextual information for improved personality trait predictions.

\vspace{0.05in}


\end{document}